\pdfoutput=1
\documentclass[11pt,a4paper]{article}
\usepackage[T1]{fontenc}
\usepackage[utf8]{inputenc}
\usepackage{lmodern}
\usepackage{amsmath,amssymb,mathtools,bm,mathrsfs}
\usepackage{microtype}
\usepackage[a4paper,margin=27mm]{geometry}
\usepackage[hidelinks]{hyperref}
\usepackage[numbers,sort&compress]{natbib}
\usepackage{enumitem}
\usepackage{booktabs}
\usepackage{graphicx}
\usepackage{float}
\usepackage{tikz}
\usetikzlibrary{arrows.meta,angles,quotes,calc}

\numberwithin{equation}{section}
\allowdisplaybreaks[3]

\newcommand{\vect}[1]{\bm{#1}}
\newcommand{\dvec}[1]{\underline{\bm{#1}}}
\newcommand{\dsc}[1]{\underline{#1}}
\newcommand{\dten}[1]{\underline{#1}}
\newcommand{\DD}{\mathbb D}
\newcommand{\RR}{\mathbb R}
\newcommand{\W}{\dvec{W}}
\newcommand{\uu}{\dvec{u}}
\newcommand{\ww}{\dvec{\omega}}

\newcommand{\Beta}{\dten{\mathcal B}}

\renewcommand{\skew}[1]{\widetilde{#1}}

\hypersetup{
  hypertexnames=false,
  pdftitle={Lie-Algebraic Bell Recurrences for Arbitrary-Order Twist Jets and Parallel-Mechanism Closure},
  pdfauthor={Daniel Condurache},
  pdfsubject={Higher-order kinematics of serial chains and parallel mechanisms},
  pdfkeywords={Bell polynomials, higher-order kinematics, dual screw theory, cylindrical joints, parallel mechanisms, twist jets, affine platform fields}
}
\title{\textbf{Lie-Algebraic Bell Recurrences for Arbitrary-Order\\
Twist Jets and Parallel-Mechanism Closure}}
\author{Daniel Condurache\\[1mm]
\small Gheorghe Asachi Technical University of Ia\c{s}i, Ia\c{s}i, Romania}
\date{}
\begin{document}
\maketitle
\begin{abstract}
This paper develops an arbitrary-order kinematic construction that links
serial propagation, parallel-mechanism closure, and rigid-platform point
fields within one dual screw framework.  A cylindrical joint is retained as
one native physical block, with revolute and prismatic joints obtained as
special cases.  For each fixed joint axis, ordinary Bell polynomials organize
the derivatives of the exponential factor; across a chain, the noncommuting
factors remain in their physical order.  Initial-frame prefix and
terminal-resolved covariant formulas then produce equivalent representations
of the serial twist jet. For a parallel mechanism, repeated Leibniz
differentiation, with joint-level derivatives organized by Bell polynomials,
yields an arbitrary-order triangular active--passive closure recurrence:
the same passive Jacobian is solved at every derivative order at a regular
configuration, while the right-hand side contains only prescribed active data
and lower-order jets.  The resulting platform twist jet is mapped exactly to
the point-independent affine invariants of the velocity, acceleration, jerk,
and snap fields. The validation is deliberately complementary: a generic
$3C$ chain tests ordered serial propagation, an $RR+RRR$ spherical wrist tests
active--passive closure, and a Hunt-type $6$-RUS mechanism with six active
revolute joints tests an independently reconstructed platform jet and its
affine fields. Independent
differentiation of the rigid motion, evaluation of the affine fields, and the
differentiated branch closures all agree through fourth order with residuals
below $10^{-12}$ in the corresponding SI units.  A separate generic $3C$ test exercises noncoplanar axes and
nonzero rotational and translational cylindrical coordinates.  The formulation
is purely kinematic and applies at
configurations where the selected active--passive partition is regular.
\end{abstract}
\noindent\textbf{Keywords:} Bell polynomials; higher-order kinematics; dual
screw theory; cylindrical joints; parallel mechanisms; twist jets; affine
platform fields.
\section{Introduction}
\label{sec:introduction}

\subsection{Motivation and scope}
\label{subsec:motivation}

Higher-order kinematics is required whenever velocity and acceleration do not
describe the motion with sufficient fidelity.  Jerk- and snap-constrained
trajectory planning, higher-order feedforward control, and repeated
differentiation of inverse dynamics all require consistent finite jets of
rigid-body motion and, ultimately, the corresponding vector fields over the
moving body \cite{Muller2014,Muller2019,Condurache2019Tensor,
Condurache2020Dual,Condurache2022,Condurache2025}.

For a serial chain, these derivatives are propagated along one ordered
sequence of joints.  A parallel mechanism adds a second problem: every branch
must generate the same platform motion.  The branchwise derivatives must
therefore be assembled in a common resolving frame and must satisfy the
successively differentiated closure equations.  After closure, the platform
twist jet must still be converted into the velocity, acceleration, jerk, and
snap fields of its material points.  A complete construction must keep these
three stages distinct:
\begin{equation}
 \text{serial propagation}
 \longrightarrow \text{parallel closure}
 \longrightarrow \text{platform fields}.
 \label{eq:C21-three-stage-map}
\end{equation}

Dual screw theory supplies the geometric language for rigid-body motion
\cite{Ball1900,Hunt1978}.  Recursive relations for velocity, acceleration,
and jerk of lower-pair chains were developed by Rico, Gallardo, and Duffy
\cite{RicoGallardoDuffy1999}; arbitrary-order and computational formulations
were subsequently established for serial and general mechanism kinematics
\cite{Lerbet1998,Muller2014,Muller2016,Muller2019,Muller2021Panda}.
Screw-theoretic analyses of parallel manipulators provide the corresponding
closed-chain setting \cite{Gallardo2016,DiGregorio2020}.

The present paper works in Condurache's native dual-vector and
orthogonal-dual-tensor convention.  The 2022 formulation supplies
arbitrary-order serial propagation and the exact operator-polynomial map from
a spatial twist jet to the affine invariants of the rigid-body fields
\cite{Condurache2022}.  The 2025 differential-transform formulation acts
instead on the configuration curve and produces the higher-order fields from
nilpotent coefficients \cite{Condurache2025}.  These procedures may agree in
their physical output, but they are not the same computational route.  This
manuscript uses the 2022 twist-jet map and does not claim a new
field-reconstruction theorem or a twist-to-transform interface.

The scope is purely kinematic.  The analysis assumes a fixed mechanism
topology and a regular active--passive partition at the evaluation
configuration.  Dynamics, compliance, contact, and the resolution of
kinematic branches at singular configurations are outside the present work.

\subsection{Native cylindrical blocks and Bell structure}
\label{subsec:native-dual-representation}

A cylindrical joint is retained as one physical block.  Its dual scalar
coordinate has independent real and dual components, representing the
coaxial rotation and translation, so no artificial intermediate body is
introduced.  Revolute and prismatic joints are recovered by suppressing one
of these two components.  For a chain with $m$ physical blocks, a cylindrical
joint is consequently counted once in $m$.

For one joint the dual axis is fixed in its incident frame.  All derivatives
of the exponential generator are scalar multiples of the same cross-product
tensor and therefore commute.  Ordinary partial and complete Bell
polynomials organize the derivatives of that individual exponential
\cite{Comtet1974}.  Generators belonging to different joints need not
commute; their factors remain in physical chain order and are combined by the
multinomial Leibniz rule.  The distinction is essential: the joint-level Bell
polynomials are commutative, whereas the chain product is order preserving.

Two resolved axis families are then used.  Prefix products propagate the
joint data from the initial frame toward the terminal body.  Suffix products
resolve the same physical data in the terminal frame and traverse the chain
in the opposite direction.  The two descriptions are not competing
kinematics; they are covariant representations of the same twist jet and are
particularly useful when several branches terminate on one platform.

\subsection{Parallel closure at arbitrary order}
\label{subsec:parallel-problem}

Let $F(q_a,q_p)=0$ denote a set of closure equations, with prescribed active
coordinates $q_a$ and unknown passive coordinates $q_p$. At derivative order
$n$, repeated Leibniz differentiation, with the joint-level derivatives
organized by Bell polynomials, collects every contribution formed from
derivatives already known at lower orders. The derivative of highest
order enters linearly.  After realification, the passive jet therefore
satisfies a triangular recurrence of the form
\begin{equation}
 J_p\,q_p^{(n+1)}=-J_a\,q_a^{(n+1)}-g_n,
 \qquad n\geq0,
 \label{eq:C21-intro-passive-recurrence}
\end{equation}
where $J_a$ and $J_p$ are the active and passive blocks of the closure
Jacobian at the current configuration, and $g_n$ depends only on lower-order
jet data.  The Bell-polynomial expression exposes the complete combinatorial
content, while the recurrence supplies its executable evaluation.

If $J_p$ is square and nonsingular, the same factorization can be reused for
all derivative orders evaluated at that instant.  This does not mean that
$J_p$ is constant along a trajectory.  Nor does the recurrence remove a
kinematic singularity: at rank loss, local uniqueness of the selected passive
coordinates is no longer guaranteed.  Differentiated closure residuals must
therefore accompany the computed jet.

Once the platform spatial twist and its derivatives are known in one fixed
frame, the exact operator polynomial $P_n[D]$, defined in
Section~\ref{sec:affine-platform-fields}, gives the point-independent
affine invariants
\begin{equation}
 \vect a_n=P_n[D]\vect v,
 \qquad
 \Phi_n=P_n[D]\widetilde{\vect\omega},
 \qquad
 \vect a_{\rho}^{[n]}=\vect a_n+\Phi_n\vect\rho.
 \label{eq:C21-intro-affine-map}
\end{equation}
Thus the twist jet through order three is sufficient for the complete
velocity, acceleration, jerk, and snap fields.  The affine stage is algebraic
and introduces no additional kinematic solve; any conditioning loss belongs
to the preceding closure stage.

\subsection{Contributions and validation strategy}
\label{subsec:contributions}

Within this scope, the paper makes the following contributions.

\begin{enumerate}[label=\arabic*.,leftmargin=8mm]
 \item It gives an explicit Bell-polynomial representation of the derivatives
 of a fixed-axis dual rotation and reduces every joint factor to the tensor
 basis generated by its unit dual axis.

 \item It derives initial-frame prefix and terminal-resolved covariant formulas for
 the arbitrary-order twist jet of an ordered native $mC$ chain, without
 splitting a cylindrical joint into fictitious consecutive joints.

 \item It assembles the branchwise terminal quantities in the common platform
 frame and formulates the active--passive closure recurrence at arbitrary
 derivative order, including the regularity condition and differentiated
 residuals.

 \item It connects the platform twist jet to the established exact affine map
 of \citet{Condurache2022} and writes the velocity, acceleration, jerk, and
 snap fields explicitly, with their vector and tensor invariants kept
 separate.

 \item It supplies three complementary internal validations. A generic $3C$
 chain tests ordered serial propagation, the $RR+RRR$ wrist tests
 active--passive closure, and a spatial test uses the Hunt-type $6$-RUS
 geometry of \citet{Gil2004}, with six active revolute joints, to test an
 independently reconstructed platform jet and its affine fields. The
 published work supplies the mechanism geometry and evaluation
 configuration; all twist-jet, jerk, snap, and affine-field results reported
 here are recomputed internally rather than fitted to its rounded numerical
 tables.
\end{enumerate}

Taken together, rather than individually, these tests cover the three stages
of the construction. The $RR+RRR$ spherical wrist retained in the closure chapters is a compact
didactic example for the active--passive recurrence.  It is not used to
validate general translational platform fields.  That role belongs to the
$6$-RUS configuration in Section~8, whose platform undergoes general spatial
motion and therefore has nonzero vector affine invariants in general.

\subsection{Organization of the paper}
\label{subsec:organization}

Section~2 defines the dual objects, the native cylindrical block, the ordered
$mC$ chain, and the jet convention.  Section~3 derives the commutative Bell
factors for one fixed dual axis while preserving joint order in the chain.
Sections~4 and 5 construct the initial-frame and terminal-resolved covariant
twist jets. Section~6 formulates the branch and platform closure equations,
their real active--passive partition, and the $RR+RRR$ wrist example.
Section~7 derives the arbitrary-order triangular Bell closure recurrence and
continues that example through snap.  Section~8 applies the exact twist-to-field
map, gives the affine velocity, acceleration, jerk, and snap fields, and
performs the internal spatial validation on the Hunt-type $6$-RUS
configuration.  Section~9 summarizes the results, states the limits of the
regular kinematic construction, and identifies the principal directions for
extension.

\section{Condurache's orthogonal-dual-tensor convention for an \texorpdfstring{$mC$}{mC} chain}
\label{sec:notation}

From this section onward the manuscript uses the convention and the chain
model of \citet[Sec.~8, Eqs.~(104)--(117)]{Condurache2022}.  The mechanism is
an ordered chain of bodies $C_0,C_1,\ldots,C_m$ connected by $m$ general
cylindrical joints.  A joint $C_k$ is one physical block; no cylindrical
joint is replaced by consecutive revolute and prismatic joints.

\subsection{Dual objects}

The ring of real dual numbers and the associated dual vectors are
\begin{equation}
 \DD=\{a+\varepsilon a_0:a,a_0\in\RR,\ \varepsilon^2=0\},
 \qquad
 \dvec a=\vect a+\varepsilon\vect a_0.
 \label{eq:C22-dual-objects}
\end{equation}
For dual vectors the tilde tensor is defined by
\begin{equation}
 \skew{\dvec a}\,\dvec b=\dvec a\times\dvec b.
 \label{eq:C22-tilde}
\end{equation}
A proper orthogonal dual tensor $\dten R$ satisfies
\begin{equation}
 \dten R^T\dten R=I,
 \qquad \det\dten R=1,
 \qquad
 \dten R\skew{\dvec a}
 =\skew{\dten R\dvec a}\,\dten R.
 \label{eq:C22-orthogonal-covariance}
\end{equation}
Equation \eqref{eq:C22-orthogonal-covariance} is the only change-of-frame
identity required below.

\subsection{One native cylindrical joint}

The relative motion of $C_k$ with respect to $C_{k-1}$ is the proper
orthogonal dual tensor
\begin{equation}
 {}^{k-1}\dten R_k
 =\exp\!\left(\dsc\theta_k\skew{{}^{k-1}\uu_k}\right)
 =I+\sin\dsc\theta_k\skew{{}^{k-1}\uu_k}
 +(1-\cos\dsc\theta_k)\skew{{}^{k-1}\uu_k}^{2},
 \label{eq:C22-relative-C-tensor}
\end{equation}
where
\begin{equation}
 \dsc\theta_k=\theta_k+\varepsilon d_k,
 \qquad
 {}^{k-1}\uu_k=\text{constant},
 \qquad
 {}^{k-1}\uu_k\mathbin{\cdot}{}^{k-1}\uu_k=1.
 \label{eq:C22-dual-angle-axis}
\end{equation}
Writing ${}^{k-1}\uu_k=\vect u_k+\varepsilon\vect u_{0k}$, the last
normalization is equivalent to
\begin{equation}
 \vect u_k\mathbin{\cdot}\vect u_k=1,
 \qquad
 \vect u_k\mathbin{\cdot}\vect u_{0k}=0,
 \label{eq:C22-dual-axis-components}
\end{equation}
so that the primal part is a unit direction and the dual part is its line
moment in the adopted convention. Thus the scalar dual rate and the relative dual twist are
\begin{equation}
 \dsc\omega_k=\dot{\dsc\theta}_k
 =\dot\theta_k+\varepsilon\dot d_k,
 \qquad
 {}^{k-1}\ww_k=\dsc\omega_k\,{}^{k-1}\uu_k.
 \label{eq:C22-relative-twist}
\end{equation}
The two components of $\dsc\theta_k$ are independent for a general $C$
joint.  Revolute, prismatic and helical pairs are only restrictions of this
single formula: respectively $d_k=0$, $\theta_k=0$, and
$d_k=h_k\theta_k$.  They are not different propagation blocks.

\subsection{Ordered chain and the two resolved axis families}

The terminal displacement is the ordered product
\begin{equation}
 {}^0\dten R_m
 ={}^0\dten R_1\,{}^1\dten R_2\cdots{}^{m-1}\dten R_m.
 \label{eq:C22-chain-product}
\end{equation}
The $k$th joint axis resolved in the initial frame is
\begin{equation}
 {}^0\uu_k
 ={}^0\dten R_{k-1}\,{}^{k-1}\uu_k,
 \qquad {}^0\dten R_0=I,
 \label{eq:C22-space-axis}
\end{equation}
which is Condurache's Eq.~(110).  The same geometric axis resolved in the
terminal frame is
\begin{equation}
 {}^m\uu_k
 ={}^m\dten R_{k-1}\,{}^{k-1}\uu_k,
 \qquad
 {}^m\dten R_{k-1}
 =({}^0\dten R_m)^T{}^0\dten R_{k-1},
 \label{eq:C22-body-axis}
\end{equation}
which is Eq.~(116).  These are two resolutions of one dual line vector, not
two differentiated coordinate curves.

In the initial frame the accumulated dual spatial twist is
\begin{equation}
 {}^0\W_k=\sum_{j=1}^k\dsc\omega_j{}^0\uu_j,
 \qquad {}^0\W_0=\dvec0.
 \label{eq:C22-prefix-twist}
\end{equation}

\subsection{Derivative convention}

Condurache's square index denotes differentiation of the relative motion
before resolution in the chosen frame:
\begin{equation}
 {}^0\ww_k^{[r]}=\dsc\omega_k^{(r)}{}^0\uu_k,
 \qquad
 {}^m\ww_k^{[r]}=\dsc\omega_k^{(r)}{}^m\uu_k.
 \label{eq:C22-square-derivative}
\end{equation}
It is not the ordinary derivative of the already resolved product
$\dsc\omega_k{}^a\uu_k(t)$.  We reserve round parentheses for ordinary time
derivatives and use $\langle n\rangle$ for the resolution of the geometric
$n$th twist derivative:
\begin{equation}
 {}^a\W_m^{\langle n\rangle}
 :=\text{components in frame $C_a$ of the geometric quantity }\W_m^{(n)}.
 \label{eq:C22-geometric-derivative-resolution}
\end{equation}
This notation prevents the moving-frame derivative error: in general
$({}^m\W_m)^{(n)}\ne{}^m\W_m^{\langle n\rangle}$.

\section{Commutative Bell polynomials for a fixed dual axis}
\label{sec:commutative-bell}

For the native cylindrical joint $C_i$, put
\begin{equation}
 \dten{A}_i:=\skew{{}^{i-1}\uu_i},
 \qquad
 Q_i(t):={}^{i-1}\dten R_i(t)
 =\exp\!\bigl(\dsc\theta_i(t)\dten{A}_i\bigr).
 \label{eq:C22-fixed-axis-generator}
\end{equation}
The dual tensor $\dten{A}_i$ is constant.  Hence all derivatives of the exponent,
$\dsc\theta_i^{(r)}\dten{A}_i$, are scalar multiples of the same tensor and commute.
No ordered or noncommutative Bell family is needed for one joint; the latter
families become relevant only when the arguments themselves do not commute
\citep{EbrahimiFard2015}.

Let $B_{n,q}$ be the ordinary partial exponential Bell polynomial and write
\[
 \mathcal B_{i;n,q}:=
 B_{n,q}\!\left(
 \dot{\dsc\theta}_i,\ddot{\dsc\theta}_i,\ldots,
 \dsc\theta_i^{(n-q+1)}
 \right).
\]
Then set
\begin{equation}
 \begin{aligned}
  \Beta_{i,0}&:=I,
  &\Beta_{i,n}&:=b^{(1)}_{i,n}\dten{A}_i+b^{(2)}_{i,n}\dten{A}_i^2
  \quad(n\geq1),\\
  b^{(1)}_{i,n}&:=
  \sum_{\substack{1\leq q\leq n\\q\ \mathrm{odd}}}
  (-1)^{(q-1)/2}\mathcal B_{i;n,q},
  &b^{(2)}_{i,n}&:=
  \sum_{\substack{2\leq q\leq n\\q\ \mathrm{even}}}
  (-1)^{q/2-1}\mathcal B_{i;n,q}.
 \end{aligned}
 \label{eq:C22-commutative-Bell-factor}
\end{equation}
Equivalently, $\Beta_{i,n}$ is the complete commutative Bell polynomial
$B_n(\dot{\dsc\theta}_i\dten{A}_i,\ldots,\dsc\theta_i^{(n)}\dten{A}_i)$.
The fixed-axis derivative is therefore
\begin{equation}
 \boxed{Q_i^{(n)}=Q_i\Beta_{i,n}=\Beta_{i,n}Q_i.}
 \label{eq:C22-relative-tensor-Bell}
\end{equation}
The reduction in \eqref{eq:C22-commutative-Bell-factor} follows because
${}^{i-1}\uu_i$ is a unit dual vector and its cross-product tensor satisfies
\[
 \dten{A}_i^3=-\dten{A}_i,
 \qquad
 \dten{A}_i^4=-\dten{A}_i^2.
\]
Consequently, the first four Bell factors reduce to the two-tensor basis
$\{\dten{A}_i,\dten{A}_i^2\}$:
\begin{align}
 \Beta_{i,1}={}&\dot{\dsc\theta}_i\dten{A}_i,\nonumber\\
 \Beta_{i,2}={}&\ddot{\dsc\theta}_i\dten{A}_i
 +(\dot{\dsc\theta}_i)^2\dten{A}_i^2,\nonumber\\
 \Beta_{i,3}={}&
 \left(\dsc\theta_i^{(3)}-(\dot{\dsc\theta}_i)^3\right)\dten{A}_i
 +3\dot{\dsc\theta}_i\ddot{\dsc\theta}_i\dten{A}_i^2,\nonumber\\
 \Beta_{i,4}={}&
 \left(\dsc\theta_i^{(4)}
 -6(\dot{\dsc\theta}_i)^2\ddot{\dsc\theta}_i\right)\dten{A}_i\nonumber\\
 &+\left(4\dot{\dsc\theta}_i\dsc\theta_i^{(3)}
 +3(\ddot{\dsc\theta}_i)^2
 -(\dot{\dsc\theta}_i)^4\right)\dten{A}_i^2.
 \label{eq:C22-relative-tensor-low-orders}
\end{align}

The inverse relative tensor has the same form with the dual angle negated:
\begin{equation}
 (Q_i^T)^{(n)}=Q_i^T\Beta^-_{i,n},
 \qquad
 \Beta^-_{i,n}:=
 \sum_{q=0}^{n}B_{n,q}\!\left(
 -\dot{\dsc\theta}_i,\ldots,-\dsc\theta_i^{(n-q+1)}
 \right)\dten{A}_i^q,
 \label{eq:C22-inverse-relative-tensor-Bell}
\end{equation}
where $B_{0,0}=1$ and $B_{n,0}=0$ for $n>0$.

Different joints need not have commuting generators.  Consequently the
relative factors remain in their physical chain order when their product is
differentiated; commutativity is used only inside each fixed-axis factor.
For $P_s:=Q_1\cdots Q_s$ the ordinary multinomial Leibniz rule gives
\begin{equation}
 P_s^{(n)}=
 \sum_{\substack{\alpha_1+\cdots+\alpha_s=n\\\alpha_j\geq0}}
 \binom{n}{\alpha_1,\ldots,\alpha_s}
 Q_1^{(\alpha_1)}\cdots Q_s^{(\alpha_s)}.
 \label{eq:C22-prefix-product-derivative}
\end{equation}
Thus every derivative required below is expressed with ordinary commutative
Bell polynomials at joint level and an order-preserving product at chain level.

\section{Higher-order kinematics of the \texorpdfstring{$mC$}{mC} chain in the initial frame}
\label{sec:initial-frame-mc}

This section is derived directly from the native cylindrical-joint model of
Section~\ref{sec:notation}.  Every joint is one $C$ block.  The only
commutativity used is the fixed-axis commutativity inside an individual joint,
already encoded by the ordinary Bell factors of
Section~\ref{sec:commutative-bell}.  Products belonging to different joints
remain in their physical order.

\subsection{Ordered prefix tensors}

Write
\begin{equation}
 \dten Q_i:={}^{i-1}\dten R_i,
 \qquad
 {}^0\dten R_k=\dten Q_1\dten Q_2\cdots\dten Q_k,
 \qquad {}^0\dten R_0=I.
 \label{eq:C24-prefix-definition}
\end{equation}
For $n\geq0$, define the $n$th ordered prefix derivative by
\begin{equation}
 {}^0\dten P_{k}^{[n]}
 :=
 \sum_{\substack{\alpha_1+\cdots+\alpha_k=n\\\alpha_j\geq0}}
 \binom{n}{\alpha_1,\ldots,\alpha_k}
 \dten Q_1\Beta_{1,\alpha_1}
 \cdots
 \dten Q_k\Beta_{k,\alpha_k},
 \qquad
 {}^0\dten P_0^{[0]}=I,
 \label{eq:C24-prefix-closed}
\end{equation}
with ${}^0\dten P_0^{[n]}=0$ for $n>0$.  Here
$\Beta_{i,0}=I$, while the factors
$\Beta_{i,n}$ for $n\geq1$ are exactly the commutative Bell factors defined in
\eqref{eq:C22-commutative-Bell-factor}.  Therefore
\begin{equation}
 \boxed{\bigl({}^0\dten R_k\bigr)^{(n)}
 ={}^0\dten P_k^{[n]}.}
 \label{eq:C24-prefix-derivative}
\end{equation}
The multinomial coefficient distributes the $n$ differentiations among the
$k$ relative tensors; it does not permute their order.

For evaluation, the same quantity obeys the forward recurrence
\begin{equation}
 {}^0\dten P_k^{[n]}
 =\sum_{r=0}^{n}\binom{n}{r}
 {}^0\dten P_{k-1}^{[n-r]}
 \dten Q_k\Beta_{k,r},
 \qquad k=1,\ldots,m.
 \label{eq:C24-prefix-recurrence}
\end{equation}
Equation~\eqref{eq:C24-prefix-recurrence} is only the binary Leibniz rule
applied to ${}^0\dten R_k={}^0\dten R_{k-1}\dten Q_k$; hence it is exactly
equivalent to the closed expression \eqref{eq:C24-prefix-closed}.

\subsection{Transported axes and their derivatives}

The $k$th constant joint axis, resolved in the initial frame, is
\begin{equation}
 {}^0\uu_k={}^0\dten R_{k-1}\,{}^{k-1}\uu_k.
 \label{eq:C24-transported-axis}
\end{equation}
Since ${}^{k-1}\uu_k$ is constant in $C_{k-1}$, its derivatives in the fixed
initial frame follow without an additional recurrence:
\begin{equation}
 \boxed{
 \bigl({}^0\uu_k\bigr)^{(r)}
 ={}^0\dten P_{k-1}^{[r]}{}^{k-1}\uu_k,
 \qquad r\geq0.}
 \label{eq:C24-axis-derivatives}
\end{equation}
Thus all time dependence of the resolved axis is carried by the ordered prefix
preceding joint $k$.

\subsection{Direct initial-frame twist jet}

The accumulated dual twist of the terminal body, resolved in the initial
frame, is
\begin{equation}
 {}^0\W_m
 =\sum_{k=1}^{m}\dot{\dsc\theta}_k\,{}^0\uu_k.
 \label{eq:C24-terminal-twist}
\end{equation}
Because $C_0$ is fixed, ordinary differentiation of these components is the
initial-frame resolution of the geometric derivative.  Applying Leibniz to
each term of \eqref{eq:C24-terminal-twist} and using
\eqref{eq:C24-axis-derivatives} gives, for every $n\geq0$,
\begin{equation}
 \boxed{
 {}^0\W_m^{\langle n\rangle}
 =\sum_{k=1}^{m}\sum_{r=0}^{n}
 \binom{n}{r}
 \dsc\theta_k^{(n-r+1)}
 {}^0\dten P_{k-1}^{[r]}
 {}^{k-1}\uu_k.}
 \label{eq:C24-initial-frame-jet}
\end{equation}
This is the direct initial-frame formula for the complete higher-order twist
jet of the $mC$ chain.  Substitution of
\eqref{eq:C24-prefix-closed} makes it an explicit expression involving only
ordinary partial Bell polynomials, joint dual-angle derivatives, constant
joint axes, and order-preserving products of relative tensors.

Equation~\eqref{eq:C24-initial-frame-jet} is used here in its autonomous
Leibniz form.  Its starting value is the initial-frame twist sum associated
with Condurache's transported axes, but no term-by-term identification with
the operator-polynomial organization of Eq.~(115) is asserted without a
separate reindexing proof.

For compactness in the low-order expansion, set
\[
 {}^0\dvec U_k^{[r]}
 :={}^0\dten P_{k-1}^{[r]}{}^{k-1}\uu_k,
 \qquad r\geq0,
\]
so that ${}^0\dvec U_k^{[0]}={}^0\uu_k$.  The first members are
\begin{align}
 {}^0\W_m^{\langle0\rangle}
 &=\sum_{k=1}^{m}\dot{\dsc\theta}_k{}^0\dvec U_k^{[0]},
 \nonumber\\
 {}^0\W_m^{\langle1\rangle}
 &=\sum_{k=1}^{m}
 \left(
 \ddot{\dsc\theta}_k{}^0\dvec U_k^{[0]}
 +\dot{\dsc\theta}_k{}^0\dvec U_k^{[1]}
 \right),
 \nonumber\\
 {}^0\W_m^{\langle2\rangle}
 &=\sum_{k=1}^{m}
 \left(
 \dsc\theta_k^{(3)}{}^0\dvec U_k^{[0]}
 +2\ddot{\dsc\theta}_k{}^0\dvec U_k^{[1]}
 +\dot{\dsc\theta}_k{}^0\dvec U_k^{[2]}
 \right),
 \nonumber\\
 {}^0\W_m^{\langle3\rangle}
 &=\sum_{k=1}^{m}
 \left(
 \dsc\theta_k^{(4)}{}^0\dvec U_k^{[0]}
 +3\dsc\theta_k^{(3)}{}^0\dvec U_k^{[1]}
 +3\ddot{\dsc\theta}_k{}^0\dvec U_k^{[2]}
 +\dot{\dsc\theta}_k{}^0\dvec U_k^{[3]}
 \right),
 \nonumber\\
 {}^0\W_m^{\langle4\rangle}
 &=\sum_{k=1}^{m}
 \left(
 \dsc\theta_k^{(5)}{}^0\dvec U_k^{[0]}
 +4\dsc\theta_k^{(4)}{}^0\dvec U_k^{[1]}
 +6\dsc\theta_k^{(3)}{}^0\dvec U_k^{[2]}
 +4\ddot{\dsc\theta}_k{}^0\dvec U_k^{[3]}
 +\dot{\dsc\theta}_k{}^0\dvec U_k^{[4]}
 \right).
 \label{eq:C24-initial-frame-low-orders}
\end{align}

No terminal-frame formula is inferred here.  In particular, this section does
not differentiate components resolved in a moving frame and does not introduce
a space--body conversion.  The result is confined to the fixed initial frame
and follows solely from the $mC$ product, the fixed dual axes, the commutative
joint-level Bell factors, and the ordinary Leibniz rule.

\section{Higher-order kinematics of the \texorpdfstring{$mC$}{mC} chain in the terminal frame}
\label{sec:terminal-frame-mc}

This section gives the terminal-frame counterpart of the initial-frame
description, directly in Condurache's convention.  The object computed is the
resolution in $C_m$ of the geometric derivative of the terminal twist.  It is
not obtained by differentiating a time-dependent array of components in the
moving frame $C_m$.

\subsection{Terminal-resolved joint axes}

For $i=1,\ldots,m$, define the proper orthogonal dual tensor from $C_{i-1}$ to
$C_m$ by
\begin{equation}
 {}^m\dten R_{i-1}
 :=({}^0\dten R_m)^T{}^0\dten R_{i-1}.
 \label{eq:C25-terminal-transport}
\end{equation}
The $i$th joint axis resolved in the terminal frame is therefore
\begin{equation}
 \boxed{{}^m\uu_i
 ={}^m\dten R_{i-1}\,{}^{i-1}\uu_i,}
 \qquad i=1,\ldots,m,
 \label{eq:C25-terminal-axis}
\end{equation}
which is precisely the axis family introduced in Condurache's Eq.~(116).
Orthogonality preserves the dual scalar product, hence
${}^m\uu_i\mathbin{\cdot}{}^m\uu_i=1$.

Put
\begin{equation}
 {}^m\dten A_i:=\skew{{}^m\uu_i}.
 \label{eq:C25-terminal-generator}
\end{equation}
Then
\begin{equation}
 ({}^m\dten A_i)^3=-{}^m\dten A_i,
 \qquad
 ({}^m\dten A_i)^4=-({}^m\dten A_i)^2.
 \label{eq:C25-terminal-generator-powers}
\end{equation}

\subsection{Joint Bell tensors resolved in \texorpdfstring{$C_m$}{Cm}}

The scalar coefficients $b^{(1)}_{i,r}$ and $b^{(2)}_{i,r}$ are those of
\eqref{eq:C22-commutative-Bell-factor}; they depend only on the derivatives of
the dual angle $\dsc\theta_i$.  Their terminal-frame tensor realization is
\begin{equation}
 {}^m\Beta_{i,0}:=I,
 \qquad
 {}^m\Beta_{i,r}
 :=b^{(1)}_{i,r}\,{}^m\dten A_i
   +b^{(2)}_{i,r}({}^m\dten A_i)^2,
 \quad r\geq1.
 \label{eq:C25-terminal-Bell-factor}
\end{equation}
Equivalently,
\begin{equation}
 {}^m\Beta_{i,r}
 ={}^m\dten R_{i-1}\Beta_{i,r}
  ({}^m\dten R_{i-1})^T.
 \label{eq:C25-terminal-Bell-covariance}
\end{equation}
Thus changing the resolution frame changes the generator but not the ordinary
commutative Bell coefficients.  Commutativity remains a one-joint property;
products belonging to distinct joints retain their chain order.

For $p=1,\ldots,m$ and $0\leq r$, define the ordered coefficient
\begin{equation}
 {}^m\dten C_{p}^{[r]}
 :=
 \sum_{\substack{\alpha_1+\cdots+\alpha_{p-1}=r\\\alpha_j\geq0}}
 \binom{r}{\alpha_1,\ldots,\alpha_{p-1}}
 {}^m\Beta_{1,\alpha_1}\cdots
 {}^m\Beta_{p-1,\alpha_{p-1}},
 \label{eq:C25-terminal-ordered-coefficient}
\end{equation}
with the empty-product convention
${}^m\dten C_{1}^{[0]}=I$ and
${}^m\dten C_{1}^{[r]}=0$ for $r>0$.
No permutation of the tensor factors is implied by the multinomial
coefficient.

\subsection{Direct terminal-frame twist jet}

The terminal formula is not an independent assumption.  It follows from the
initial-frame expression by the following covariance argument.

\paragraph{Lemma (initial-to-terminal covariance).}
For every order $n$, resolving Eq.~\eqref{eq:C24-initial-frame-jet} in $C_m$
gives the terminal expression below term by term.

\emph{Proof.}
Fix $p$ and a multi-index
$\boldsymbol\alpha=(\alpha_1,\ldots,\alpha_{p-1})$. Put
$T_i:={}^m\dten R_{i-1}=({}^0\dten R_m)^T{}^0\dten R_{i-1}$.
Since ${}^0\dten R_i={}^0\dten R_{i-1}Q_i$, one has
$T_{i+1}=T_iQ_i$. Moreover,
\begin{equation}
 {}^m\Beta_{i,\alpha_i}=T_i\Beta_{i,\alpha_i}T_i^T,
 \qquad
 {}^m\uu_p=T_p{}^{p-1}\uu_p.
 \label{eq:C25-proof-conjugates}
\end{equation}
Because $Q_i\Beta_{i,\alpha_i}=\Beta_{i,\alpha_i}Q_i$ for factors of the
same joint, successive insertion of $T_i^TT_i=I$ gives
\begin{align}
 &({}^0\dten R_m)^T
 Q_1\Beta_{1,\alpha_1}\cdots
 Q_{p-1}\Beta_{p-1,\alpha_{p-1}}{}^{p-1}\uu_p
 \nonumber\\
 &\quad={}^{m}\Beta_{1,\alpha_1}\cdots
 {}^{m}\Beta_{p-1,\alpha_{p-1}}{}^m\uu_p.
 \label{eq:C25-proof-ordered-identity}
\end{align}
No factors belonging to distinct joints have been commuted. Summing
Eq.~\eqref{eq:C25-proof-ordered-identity} over all multi-indices of total
degree $r$, with their multinomial coefficients, produces
${}^m\dten C_p^{[r]}{}^m\uu_p$ and proves the claim. \hfill$\square$

Thus Condurache's Eq.~(117) takes the following form in the notation of this
manuscript:
\begin{equation}
 \boxed{
 {}^m\W_m^{\langle n\rangle}
 =\sum_{p=1}^{m}\sum_{r=0}^{n}
 \binom{n}{r}
 {}^m\dten C_{p}^{[r]}
 \dsc\theta_p^{(n-r+1)}{}^m\uu_p,
 \qquad n\geq0.}
 \label{eq:C25-terminal-frame-jet}
\end{equation}
The square-index derivatives of Section~\ref{sec:notation} appear here as
$\dsc\theta_p^{(s+1)}{}^m\uu_p$.  The tensors
${}^m\dten C_p^{[r]}$ collect the preceding joint contributions in their
physical order.  Formula \eqref{eq:C25-terminal-frame-jet} is evaluated at the
same instant as the terminal-resolved axes; it does not differentiate those
moving-frame coordinates.

For compactness, set
\begin{equation}
 {}^m\dvec V_{p}^{[r]}
 :={}^m\dten C_p^{[r]}{}^m\uu_p.
 \label{eq:C25-terminal-V-definition}
\end{equation}
The first five members of the geometric jet are then
\begin{align}
 {}^m\W_m^{\langle0\rangle}
 &=\sum_{p=1}^{m}
 \dot{\dsc\theta}_p{}^m\dvec V_p^{[0]},
 \nonumber\\
 {}^m\W_m^{\langle1\rangle}
 &=\sum_{p=1}^{m}
 \left(
 \ddot{\dsc\theta}_p{}^m\dvec V_p^{[0]}
 +\dot{\dsc\theta}_p{}^m\dvec V_p^{[1]}
 \right),
 \nonumber\\
 {}^m\W_m^{\langle2\rangle}
 &=\sum_{p=1}^{m}
 \left(
 \dsc\theta_p^{(3)}{}^m\dvec V_p^{[0]}
 +2\ddot{\dsc\theta}_p{}^m\dvec V_p^{[1]}
 +\dot{\dsc\theta}_p{}^m\dvec V_p^{[2]}
 \right),
 \nonumber\\
 {}^m\W_m^{\langle3\rangle}
 &=\sum_{p=1}^{m}
 \left(
 \dsc\theta_p^{(4)}{}^m\dvec V_p^{[0]}
 +3\dsc\theta_p^{(3)}{}^m\dvec V_p^{[1]}
 +3\ddot{\dsc\theta}_p{}^m\dvec V_p^{[2]}
 +\dot{\dsc\theta}_p{}^m\dvec V_p^{[3]}
 \right),
 \nonumber\\
 {}^m\W_m^{\langle4\rangle}
 &=\sum_{p=1}^{m}
 \left(
 \dsc\theta_p^{(5)}{}^m\dvec V_p^{[0]}
 +4\dsc\theta_p^{(4)}{}^m\dvec V_p^{[1]}
 +6\dsc\theta_p^{(3)}{}^m\dvec V_p^{[2]}
 +4\ddot{\dsc\theta}_p{}^m\dvec V_p^{[3]}
 +\dot{\dsc\theta}_p{}^m\dvec V_p^{[4]}
 \right).
 \label{eq:C25-terminal-low-orders}
\end{align}

The distinction between geometric differentiation and coordinate
differentiation is essential.  In general,
\begin{equation}
 \left({}^m\W_m\right)^{(n)}
 \neq{}^m\W_m^{\langle n\rangle},
 \qquad n\geq1.
 \label{eq:C25-moving-frame-warning}
\end{equation}
Accordingly, no descending suffix state, no differentiated space--body
identity, and no extra transport recurrence are introduced in this section.
The result is the direct terminal-resolved covariant representation supplied by
Eqs.~(116)--(117) of \citet{Condurache2022}.

\subsection{Direct generic cylindrical-chain check}

A reproducible test uses a generic $3C$ chain with the noncoplanar local axes
\begin{equation}
 (1,0,0)^T,\qquad (0,1,1)^T/\sqrt2,\qquad
 (1,-1,1)^T/\sqrt3.
 \label{eq:C25-generic-3c-axes}
\end{equation}
Independent polynomial laws are assigned to every $\theta_i(t)$ and $d_i(t)$;
all six rates are nonzero at $t_0=2/7\,\mathrm s$.  The supplied script
\texttt{validate\_generic\_3c.py} differentiates each native $C$ factor by a
complex Cauchy-coefficient calculation and forms the ordered multinomial
prefix sums through order four.  These are compared with an independent
differentiation of the complete product.  The largest absolute discrepancy is
below $10^{-9}$, while initial/terminal twist covariance closes below
$10^{-15}$.  This test activates the translational component of the
cylindrical block and checks factor order for noncommuting joint axes.

\section{Dual model and closure equations of a parallel mechanism}
\label{sec:parallel-closure}

The serial-chain results of Sections~\ref{sec:notation}--
\ref{sec:terminal-frame-mc} are now used as branch models for a parallel
mechanism.  The construction is the dual-tensor counterpart of the standard
open-chain decomposition of a parallel robot: every branch is evaluated as an
open serial chain, while the common platform supplies the closure equations
\citep[Ch.~7]{LynchPark2017}.  This section establishes the configuration and
first-order closure only.  Higher-order differentiation of that closure is
deferred to the next section of the manuscript.

\subsection{Branches, attachments and coordinates}

Let a fixed base $C_0$ and a moving platform $C_P$ be connected by $L\geq2$
serial branches.  Branch $\ell$ contains $m_\ell$ native cylindrical joints
and the ordered bodies
\[
 C_{\ell,0},C_{\ell,1},\ldots,C_{\ell,m_\ell},
 \qquad \ell=1,\ldots,L.
\]
The base attachment and platform attachment are represented by the constant
proper orthogonal dual tensors
\begin{equation}
 {}^0\dten R_{\ell,0},
 \qquad
 {}^{\ell,m_\ell}\dten R_P,
 \label{eq:C26-fixed-attachments}
\end{equation}
respectively.  Introducing both tensors prevents the terminal body of a
branch from being identified implicitly with the platform frame.

The $i$th joint of branch $\ell$ has the dual coordinate and fixed native axis
\begin{equation}
 \dsc\theta_{\ell i}=\theta_{\ell i}+\varepsilon d_{\ell i},
 \qquad
 {}^{\ell,i-1}\uu_{\ell i}\mathbin{\cdot}
 {}^{\ell,i-1}\uu_{\ell i}=1.
 \label{eq:C26-branch-coordinate-axis}
\end{equation}
Its relative tensor is the specialization of
\eqref{eq:C22-relative-C-tensor},
\begin{equation}
 {}^{\ell,i-1}\dten R_{\ell,i}
 =\exp\!\left(
 \dsc\theta_{\ell i}\skew{{}^{\ell,i-1}\uu_{\ell i}}
 \right).
 \label{eq:C26-branch-joint-tensor}
\end{equation}
Write
\begin{equation}
 \dvec q_\ell
 :=(\dsc\theta_{\ell1},\ldots,
       \dsc\theta_{\ell m_\ell})^T,
 \qquad
 \dvec q:=(\dvec q_1^T,\ldots,\dvec q_L^T)^T.
 \label{eq:C26-branch-coordinates}
\end{equation}

\subsection{Branch forward kinematics and position closure}

Regarded as an open chain, branch $\ell$ predicts the platform displacement
\begin{equation}
 {}^0\dten R_P^{(\ell)}(\dvec q_\ell)
 ={}^0\dten R_{\ell,0}
 \left(
 \prod_{i=1}^{m_\ell}{}^{\ell,i-1}\dten R_{\ell,i}
 \right)
 {}^{\ell,m_\ell}\dten R_P,
 \label{eq:C26-branch-forward-map}
\end{equation}
where the product is ordered by increasing $i$.  The configuration closure is
\begin{equation}
 \boxed{
 {}^0\dten R_P
 ={}^0\dten R_P^{(1)}(\dvec q_1)
 =\cdots=
 {}^0\dten R_P^{(L)}(\dvec q_L).}
 \label{eq:C26-position-closure}
\end{equation}
Equivalently, the platform variable can be eliminated by selecting branch
$1$ as a reference and imposing
\begin{equation}
 \left({}^0\dten R_P^{(1)}\right)^T
 {}^0\dten R_P^{(\ell)}=I,
 \qquad \ell=2,\ldots,L.
 \label{eq:C26-relative-position-closure}
\end{equation}
The reference branch is only an algebraic choice; it has no distinguished
physical role.

\subsection{Platform-resolved branch axes and twist compatibility}

For each branch let
\begin{equation}
 {}^P\dten R_{\ell,m_\ell}
 :=\left({}^{\ell,m_\ell}\dten R_P\right)^T.
 \label{eq:C26-terminal-to-platform}
\end{equation}
The axis of joint $(\ell,i)$ resolved directly in the common platform frame
is
\begin{equation}
 {}^P\uu_{\ell i}
 :={}^P\dten R_{\ell,m_\ell}
 {}^{\ell,m_\ell}\uu_{\ell i},
 \label{eq:C26-platform-axis}
\end{equation}
where ${}^{\ell,m_\ell}\uu_{\ell i}$ is obtained from
\eqref{eq:C25-terminal-axis} within branch $\ell$.  Thus the constant mounting
tensor changes only the resolution of the terminal formula.

Define the dual branch Jacobian matrix by its ordered columns,
\begin{equation}
 {}^P\dten J_\ell(\dvec q_\ell)
 :=\begin{bmatrix}
 {}^P\uu_{\ell1}&\cdots&{}^P\uu_{\ell m_\ell}
 \end{bmatrix}.
 \label{eq:C26-branch-jacobian}
\end{equation}
The order-zero specialization of \eqref{eq:C25-terminal-frame-jet} gives the
platform twist predicted by branch $\ell$:
\begin{equation}
 {}^P\W_P^{(\ell)}
 ={}^P\dten J_\ell\dot{\dvec q}_\ell
 =\sum_{i=1}^{m_\ell}
 \dot{\dsc\theta}_{\ell i}\,{}^P\uu_{\ell i}.
 \label{eq:C26-branch-platform-twist}
\end{equation}
Since all terminal attachments belong to the same rigid platform, admissible
motions satisfy
\begin{equation}
 \boxed{
 {}^P\W_P^{(1)}=\cdots={}^P\W_P^{(L)}={}^P\W_P.}
 \label{eq:C26-twist-compatibility}
\end{equation}
This equality compares geometric twists already resolved in the same frame;
it does not differentiate a coordinate array in the moving frame $C_P$.

\subsection{Global first-order closure matrix}

Eliminating ${}^P\W_P$ from
\eqref{eq:C26-twist-compatibility} gives $L-1$ dual-vector equations.  With
the branch ordering of \eqref{eq:C26-branch-coordinates}, define
\begin{equation}
 \dten H(\dvec q):=
 \begin{bmatrix}
 -{}^P\dten J_1&{}^P\dten J_2&0&\cdots&0\\
 -{}^P\dten J_1&0&{}^P\dten J_3&\cdots&0\\
 \vdots&\vdots&\vdots&\ddots&\vdots\\
 -{}^P\dten J_1&0&0&\cdots&{}^P\dten J_L
 \end{bmatrix}.
 \label{eq:C26-global-H}
\end{equation}
Then the stacked first-order closure, containing all comparisons with the
reference branch but not necessarily a minimal independent row set, is
\begin{equation}
 \boxed{\dten H(\dvec q)\dot{\dvec q}=\dvec0.}
 \label{eq:C26-global-velocity-closure}
\end{equation}
Choosing another reference branch premultiplies and recombines the block
equations but leaves their common nullspace unchanged.

\subsection{Realification and the active--passive partition}

For a general $C$ joint, the real variables $\theta_{\ell i}$ and
$d_{\ell i}$ are independent.  Consequently, rank and regularity are stated
after realification rather than by treating one dual coordinate as one real
degree of freedom.  Define
\begin{equation}
 \mathfrak r(\vect a+\varepsilon\vect a_0)
 :=\begin{bmatrix}\vect a\\\vect a_0\end{bmatrix},
 \qquad \mathfrak r:\DD^3\longrightarrow\RR^6.
 \label{eq:C26-realification-vector}
\end{equation}
If a dual matrix is
$\dten M=\vect M+\varepsilon\vect M_0$, its real action is represented by
\begin{equation}
 \widehat M
 :=\begin{bmatrix}
 \vect M&0\\
 \vect M_0&\vect M
 \end{bmatrix},
 \qquad
 \mathfrak r(\dten M\dvec x)
 =\widehat M\,\mathfrak r(\dvec x).
 \label{eq:C26-realification-matrix}
\end{equation}
For an unambiguous global construction, replace every dual column
$\dvec h_j=\vect a_j+\varepsilon\vect a_{0j}$ of $\dten H$ by the real
$6\times2$ block
\begin{equation}
 \mathscr R(\dvec h_j)
 :=\begin{bmatrix}
 \vect a_j&0\\
 \vect a_{0j}&\vect a_j
 \end{bmatrix}.
 \label{eq:C26-realification-column}
\end{equation}
The blocks are placed in the same order as the intercalated coordinate pairs,
and the six rows of each branch comparison are ordered as primal components
followed by dual components. Applying this rule to
\eqref{eq:C26-global-H} defines, without an implicit permutation, the ordinary
real constraint
\begin{equation}
 \widehat H(\vect x)\dot{\vect x}=0,
 \qquad
 \vect x
 :=(\theta_{11},d_{11},\ldots,
       \theta_{L m_L},d_{L m_L})^T,
 \label{eq:C26-real-velocity-closure}
\end{equation}
with $\widehat H\in\RR^{6(L-1)\times2M}$ and
$M:=\sum_{\ell=1}^L m_\ell$.

Let
\begin{equation}
 \varrho_H:=\operatorname{rank}\widehat H(\vect x).
 \label{eq:C26-closure-rank}
\end{equation}
Choose, locally at the configuration under consideration, a row-selection matrix
$S\in\RR^{\varrho_H\times6(L-1)}$ such that
\begin{equation}
 \overline H:=S\widehat H
 \quad\hbox{has row rank }\varrho_H.
 \label{eq:C26-independent-closure}
\end{equation}
Thus $\widehat H$ is the complete stacked comparison system, whereas
$\overline H$ is a local independent system with the same admissible tangent
space.

Partition the real coordinates into prescribed active coordinates
$\vect x_a$ and dependent passive coordinates $\vect x_p$. The corresponding
column partition of the independent system is
\begin{equation}
 \overline H_a\dot{\vect x}_a
 +\overline H_p\dot{\vect x}_p=0.
 \label{eq:C26-active-passive-closure}
\end{equation}
For the square independent formulation one must choose
$\dim\vect x_p=\varrho_H$ and therefore
$\dim\vect x_a=2M-\varrho_H$. At a regular configuration, the passive coordinates can then be chosen so that
$\overline H_p$ is square and nonsingular. The passive rates
are then uniquely determined by
\begin{equation}
 \boxed{
 \dot{\vect x}_p
 =-\overline H_p^{-1}\overline H_a\dot{\vect x}_a.}
 \label{eq:C26-passive-rates}
\end{equation}
If instead a redundant rectangular system is retained and its passive block
$\widehat H_p$ has full column rank, a solution exists precisely when
\begin{equation}
 (I-\widehat H_p\widehat H_p^{\dagger})
 \widehat H_a\dot{\vect x}_a=0.
 \label{eq:C26-rectangular-compatibility}
\end{equation}
Under this compatibility condition the unique passive rate is
\begin{equation}
 \dot{\vect x}_p
 =-\widehat H_p^{\dagger}\widehat H_a\dot{\vect x}_a.
 \label{eq:C26-rectangular-passive-rates}
\end{equation}
The inverse formula \eqref{eq:C26-passive-rates} is reserved for the square
independent system.

Both the selector $S$ and this coordinate partition are local: they remain
valid only while the selected rows stay independent and $\overline H_p$
stays nonsingular. Loss of rank of $\overline H_p$ is a singularity of this local passive-rate
resolution.  Loss of rank of the complete $\widehat H$ describes a dependence
among the selected closure equations.  These two rank tests must not be
identified without regard to the chosen actuation partition.

For mechanisms containing restricted lower pairs, the reduction is made
before the active--passive partition.  A revolute joint has
$d_{\ell i}\equiv0$, so its identically zero translational coordinate and the
corresponding column are removed from $\vect x$.  For the spherical wrist
below, the dual closure components at the common centre $O$ vanish
identically; the selector $S$ therefore retains the three independent primal
rows only.  This converts the general realified system to the displayed
$3\times3$ passive system.  The same rule removes nonexistent $d$ coordinates
for the actuated revolutes of the $6$-RUS example.

\subsection{Worked example: the \texorpdfstring{$RR+RRR$}{RR+RRR} spherical parallel wrist}

We now specialize the closure construction to the two-degree-of-freedom
parallel wrist developed in \citet[Ch.~9, pp.~205--217]{Gallardo2016}.  This is
not a scalar surrogate: it is a spatial parallel mechanism with two different
branches.  All five revolute axes meet at $O$, adjacent axes are orthogonal,
and the platform therefore has a spherical motion with zero translational
twist at $O$.  The first branch is $RR$, the second is $RRR$, and the first
joint of each branch is actuated.

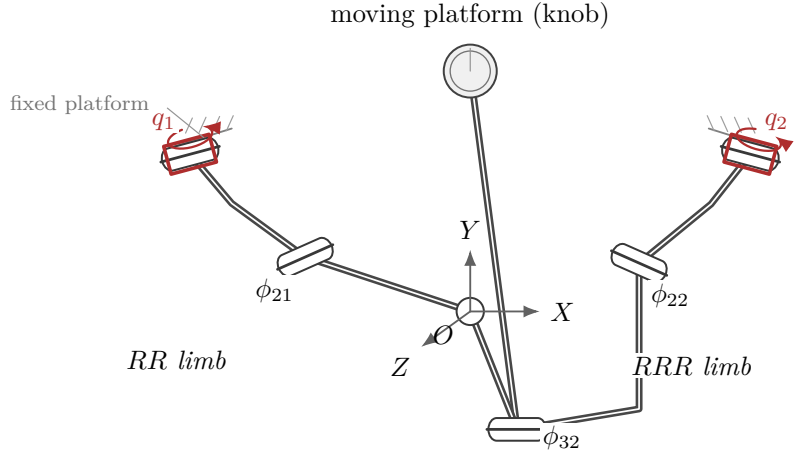
\begin{figure}[tb]
\centering
\begin{tikzpicture}[x=1cm,y=1cm,>=Latex,line cap=round,line join=round,
  link/.style={draw=black!72,line width=2.6pt},
  edge/.style={draw=white,line width=.65pt},
  shaft/.style={draw=black!78,line width=1.05pt},
  axis/.style={->,draw=black!62,line width=.7pt},
  leader/.style={draw=black!42,line width=.45pt},
  tag/.style={font=\small,fill=white,inner sep=1.2pt}]
  \definecolor{actred}{RGB}{174,45,45}

  \coordinate (O)  at (0,0);
  \coordinate (C)  at (0,3.18);
  \coordinate (A1) at (-3.70,2.08);
  \coordinate (P1) at (-2.18,.72);
  \coordinate (A2) at ( 3.70,2.08);
  \coordinate (P3) at ( 2.23,.64);
  \coordinate (P2) at ( .62,-1.56);

  \foreach \x/\sgn in {-3.70/-1,3.70/1}{
    \draw[black!50,line width=.7pt] (\x-.55*\sgn,2.42)--(\x+.28*\sgn,2.16);
    \foreach \d in {0,.18,.36,.54}{
      \draw[black!38,line width=.45pt]
        (\x-.48*\sgn+\d*\sgn,2.39-.057*\d)--++(-.09*\sgn,.20);
    }
  }

  \draw[link] (P2)--(C); \draw[edge] (P2)--(C);
  \filldraw[fill=black!6,draw=black!75,line width=.75pt] (C) circle (3.5mm);
  \draw[black!55,line width=.5pt] (C) circle (2.65mm);

  \draw[link] (A1)--(-3.15,1.43)--(P1)--(O); \draw[edge] (A1)--(-3.15,1.43)--(P1)--(O);
  \draw[link] (A2)--(3.18,1.42)--(P3)--(2.23,-1.29)--(P2)--(O);
  \draw[edge] (A2)--(3.18,1.42)--(P3)--(2.23,-1.29)--(P2)--(O);

  \foreach \P/\ang in {A1/15,P1/24,A2/-15,P3/-24,P2/0}{
    \begin{scope}[shift={(\P)},rotate=\ang]
      \filldraw[fill=white,draw=black!78,line width=.75pt]
        (-.27,-.15) arc[start angle=270,end angle=90,x radius=.11,y radius=.15] --
        (.27,.15) arc[start angle=90,end angle=-90,x radius=.11,y radius=.15] -- cycle;
      \draw[shaft] (-.40,0)--(.40,0);
    \end{scope}
  }

  \foreach \P/\ang in {A1/15,A2/-15}{
    \begin{scope}[shift={(\P)},rotate=\ang]
      \draw[actred,line width=1.35pt] (-.31,-.19) rectangle (.31,.19);
      \draw[->,actred,line width=.85pt] (-.02,.34) arc[start angle=105,end angle=410,x radius=.30,y radius=.13];
    \end{scope}
  }

  \filldraw[fill=white,draw=black!78,line width=.75pt] (O) circle (1.8mm);
  \draw[axis] (O)--(.92,0) node[right,font=\small] {$X$};
  \draw[axis] (O)--(0,.82) node[above,font=\small] {$Y$};
  \draw[axis] (O)--(-.65,-.47) node[below left,font=\small] {$Z$};
  \node[font=\small,anchor=north east] at (-.08,-.06) {$O$};

  \node[tag,text=actred,anchor=south east] at (-3.84,2.35) {$q_1$};
  \node[tag,text=actred,anchor=south west] at ( 3.84,2.35) {$q_2$};
  \node[tag,anchor=north east] at (-2.28,.49) {$\phi_{21}$};
  \node[tag,anchor=north west] at ( 2.34,.42) {$\phi_{22}$};
  \node[tag,anchor=west] at (.91,-1.67) {$\phi_{32}$};
  \node[tag,font=\small\itshape,anchor=east] at (-3.18,-.64) {$RR$ limb};
  \node[tag,font=\small\itshape,anchor=west] at ( 2.08,-.72) {$RRR$ limb};

  \node[font=\small,anchor=south] at (0,3.61) {moving platform (knob)};
  \draw[leader] (0,3.48)--(C);
  \node[font=\scriptsize,text=black!55,anchor=east] at (-4.10,2.74)
    {fixed platform};
  \draw[leader] (-4.00,2.69)--(-3.52,2.30);
\end{tikzpicture}

\caption{Kinematic schematic of the $RR+RRR$ spherical wrist, adapted from
\citet[Figs.~9.1--9.2]{Gallardo2016}.  The axonometric view preserves the limb
topology and the concurrency of all five revolute axes at $O$; red collars and
arrows identify the two actuated pairs.  The view is not a dimensioned
manufacturing drawing.}
\label{fig:C26-wrist}
\end{figure}

Let $r_w$ be the wrist radius and let $q_1,q_2$ be the active coordinates.
Following the notation of the source, the compatible position is available in
closed form.  In the fixed frame $XYZ$,
\begin{align}
 \vect p_1&=r_w(0,-\sin q_1,\cos q_1)^T,
 &\vect p_3&=r_w(\cos q_2,\sin q_2,0)^T,\nonumber\\
 \vect c&=\lambda(-\tan q_2,1,\tan q_1)^T,
 &\lambda&=\frac{r_w}{\sqrt{1+\tan^2q_1+\tan^2q_2}}.
 \label{eq:C26-wrist-position-data}
\end{align}
These are Eqs.~(9.1), (9.4), and (9.5) of \citet{Gallardo2016}.  They define
the three moving axes needed below,
\begin{equation}
 \vect u_1=\vect p_1/r_w,\qquad
 \vect u_2=\vect p_3/r_w,\qquad
 \vect u_C=\vect c/r_w.
 \label{eq:C26-wrist-passive-axes}
\end{equation}

The data of the first-order problem are the compatible configuration and the
two active rates.  The unknowns are the three passive rates
$\dot{\vect x}_p=(\omega_{21},\omega_{22},\omega_{32})^T$ and the platform
angular velocity $\vect\omega$.  Equality of the two branch predictions is
\begin{equation}
 \underbrace{\begin{bmatrix}\vect e_X&-\vect e_Z\end{bmatrix}}_{\overline H_a}
 \begin{bmatrix}\dot q_1\\\dot q_2\end{bmatrix}
 +
 \underbrace{\begin{bmatrix}\vect u_1&-\vect u_2&-\vect u_C\end{bmatrix}}_{\overline H_p}
 \begin{bmatrix}\omega_{21}\\\omega_{22}\\\omega_{32}\end{bmatrix}=0.
 \label{eq:C26-wrist-independent-system}
\end{equation}
After solving it, the common platform twist is reconstructed from either
branch:
\begin{equation}
 \vect\omega=\dot q_1\vect e_X+\omega_{21}\vect u_1
 =\dot q_2\vect e_Z+\omega_{22}\vect u_2+\omega_{32}\vect u_C,
 \qquad \vect v_O=0.
 \label{eq:C26-wrist-platform-reconstruction}
\end{equation}

For a reproducible numerical instance, let $L$ denote the length unit and
introduce the dimensionless phase $\tau=\Omega t$.  The trajectory of
\citet[Example~9.3]{Gallardo2016} is then written in dimensionally explicit
form as
\begin{equation}
 r_w=0.1L,\qquad q_1(t)=\frac{\pi}{2}\sin^2\tau\ \mathrm{rad},\qquad
 q_2(t)=\frac{\pi}{2}\sin\tau\cos\tau\ \mathrm{rad},
 \qquad \tau=\Omega t,
 \label{eq:C26-wrist-source-trajectory}
\end{equation}
where the numerical evaluation uses $\Omega=1\ \mathrm{s}^{-1}$.  Hence
$\tau_0=\pi/6$ corresponds to $t_0=\pi/(6\Omega)=\pi/6\ \mathrm{s}$, and
\begin{equation}
 \vect x_a(t_0)=(0.392699,\,0.680175)^T\ \mathrm{rad},\qquad
 \dot{\vect x}_a(t_0)=(1.360350,\,0.785398)^T\ \mathrm{rad/s}.
 \label{eq:C26-wrist-active-data}
\end{equation}
The passive block at this configuration is
\begin{equation}
 \overline H_p=
 \begin{bmatrix}
 0&-0.777463&0.598652\\
 -0.382683&-0.628929&-0.740036\\
 0.923880&0&-0.306533
 \end{bmatrix},\qquad
 \det\overline H_p=0.970604\ne0.
 \label{eq:C26-wrist-Hp-numeric}
\end{equation}
Consequently, Eq.~\eqref{eq:C26-passive-rates} gives
\begin{equation}
 \boxed{\dot{\vect x}_p(t_0)
 =(0.500030,\,0.937274,\,-1.055127)^T\ \mathrm{rad/s}.}
 \label{eq:C26-wrist-passive-result}
\end{equation}
The reconstructed platform twist is
\begin{equation}
 \boxed{{}^0\vect W_P(t_0)=
\bigl(\vect\omega;\vect v_O\bigr)
=\bigl((1.360350,-0.191353,0.461967)^T\ \mathrm{rad/s};
 (0,0,0)^T\ L/\mathrm{s}\bigr),}
 \label{eq:C26-wrist-platform-result}
\end{equation}
where angular components precede the vanishing linear components and $L$ is
the same length unit used for $r_w$.  Substitution
in the two expressions of \eqref{eq:C26-wrist-platform-reconstruction} gives
the same $\vect\omega$ to the displayed precision. Substitution of the
six-decimal values printed above gives
$\|\overline H_a\dot{\vect x}_a+\overline H_p\dot{\vect x}_p\|_2<10^{-6}$;
with the unrounded internal values used by the validation script, the residual
is below $10^{-12}$.
The section's abstract map is therefore realized here as a genuine spatial
parallel-wrist computation:
\begin{equation}
 (\vect x(t_0),\dot{\vect x}_a(t_0))
 \longmapsto (\dot{\vect x}_p(t_0),{}^0\vect W_P(t_0)).
 \label{eq:C26-wrist-input-output-map}
\end{equation}

\subsection{Result and boundary of the section}

Equations \eqref{eq:C26-position-closure} and
\eqref{eq:C26-global-velocity-closure} are the configuration and first-order
closure of the parallel mechanism.  Their construction is branchwise: the
serial formulas produce platform-resolved twists, and equality at the common
platform couples the branches.  The fixed attachment tensors retain the
specific geometry of the loops that converge to the platform.

No higher-order closure term has been introduced here.  The next section will
differentiate \eqref{eq:C26-twist-compatibility} geometrically and use the Bell
factors of each serial branch to show that, at every order, the highest passive
derivative is multiplied by the same instantaneous matrix
$\widehat H_p$ while lower-order derivatives enter only through a known
residual.

\section{Bell recurrence for arbitrary-order parallel closure}
\label{sec:parallel-bell-recurrence}

This section extends the first-order closure of
Section~\ref{sec:parallel-closure} to the complete geometric twist jet.  The
construction is performed branchwise in the common platform frame and uses
only the terminal serial-chain formula of
Section~\ref{sec:terminal-frame-mc}.  At a fixed regular configuration, the
same real closure matrix that determines the passive rates also multiplies
the highest passive derivative at every subsequent order.

\subsection{Branchwise separation of the highest derivative}

For joint $p$ of branch $\ell$, transport the ordered coefficient and vector
of \eqref{eq:C25-terminal-ordered-coefficient}--
\eqref{eq:C25-terminal-V-definition} through the constant terminal mounting
and denote the result by
\begin{equation}
 {}^P\dten C_{\ell p}^{[r]},
 \qquad
 {}^P\dvec V_{\ell p}^{[r]}
 :={}^P\dten C_{\ell p}^{[r]}{}^P\uu_{\ell p},
 \qquad
 {}^P\dvec V_{\ell p}^{[0]}={}^P\uu_{\ell p}.
 \label{eq:C27-platform-branch-V}
\end{equation}
The terminal formula \eqref{eq:C25-terminal-frame-jet}, applied to branch
$\ell$, is
\begin{equation}
 {}^P\W_P^{(\ell,\langle n\rangle)}
 =\sum_{p=1}^{m_\ell}\sum_{s=0}^{n}
 \binom ns{}^P\dvec V_{\ell p}^{[s]}
 \dsc\theta_{\ell p}^{(n-s+1)}.
 \label{eq:C27-branch-jet}
\end{equation}
Separate the term $s=0$ and define the known lower-order residual
\begin{equation}
 {}^P\dvec\rho_{\ell,n}
 :=\sum_{p=1}^{m_\ell}\sum_{s=1}^{n}
 \binom ns{}^P\dvec V_{\ell p}^{[s]}
 \dsc\theta_{\ell p}^{(n-s+1)},
 \qquad {}^P\dvec\rho_{\ell,0}:=\dvec0.
 \label{eq:C27-branch-residual}
\end{equation}
Hence
\begin{equation}
 \boxed{
 {}^P\W_P^{(\ell,\langle n\rangle)}
 ={}^P\dten J_\ell\dvec q_\ell^{(n+1)}
  +{}^P\dvec\rho_{\ell,n}.}
 \label{eq:C27-highest-derivative-split}
\end{equation}
This repeated-Leibniz organization is consistent with the standard
Bell-polynomial form of composed derivatives \citep{Schumann2019}. Every Bell factor in the
residual involves derivatives of order at most $n$;
therefore the only derivative of order $n+1$ in
\eqref{eq:C27-highest-derivative-split} is the displayed linear term.

\subsection{Arbitrary-order closure theorem}

Define the stacked dual residual difference
\begin{equation}
 \dvec g_n:=
 \begin{bmatrix}
 {}^P\dvec\rho_{2,n}-{}^P\dvec\rho_{1,n}\\
 \vdots\\
 {}^P\dvec\rho_{L,n}-{}^P\dvec\rho_{1,n}
 \end{bmatrix}.
 \label{eq:C27-stacked-residual}
\end{equation}

\paragraph{Theorem (triangular Bell closure).}
Assume that the branch configurations satisfy the position closure
\eqref{eq:C26-position-closure}.  Then, for every $n\geq0$, the order-$n$
geometric twist derivatives of the branches agree if and only if
\begin{equation}
 \boxed{
 \dten H(\dvec q)\dvec q^{(n+1)}=-\dvec g_n.}
 \label{eq:C27-dual-order-n-closure}
\end{equation}

\paragraph{Proof.}
For each $\ell=2,\ldots,L$, subtract the branch-$1$ instance of
\eqref{eq:C27-highest-derivative-split} from the branch-$\ell$ instance.
The highest-order terms give the corresponding block row
$[-{}^P\dten J_1\ \ {}^P\dten J_\ell]$ of $\dten H$, while the remaining
terms give
${}^P\dvec\rho_{\ell,n}-{}^P\dvec\rho_{1,n}$.  Stacking the differences
proves \eqref{eq:C27-dual-order-n-closure}.  Reversing the subtraction proves
the converse. \hfill$\square$

At $n=0$, $\dvec g_0=\dvec0$, so the theorem reduces exactly to
\eqref{eq:C26-global-velocity-closure}.  For $n\geq1$, all new nonlinear terms
are confined to the known right-hand side.

\subsection{Real active--passive recurrence}

Realify $\dvec g_n$ with the same branchwise primal--dual row ordering used
for $\widehat H$ and denote the result by $\widehat{\vect g}_n$. Applying the
row selection $S$ of \eqref{eq:C26-independent-closure} gives
\begin{equation}
 \overline H\vect x^{(n+1)}
 =-S\widehat{\vect g}_n.
 \label{eq:C27-independent-real-closure}
\end{equation}
For the active--passive partition fixed in Section~6,
\begin{equation}
 \boxed{
 \overline H_p\vect x_p^{(n+1)}
 =-\overline H_a\vect x_a^{(n+1)}
  -S\widehat{\vect g}_n.}
 \label{eq:C27-passive-recurrence}
\end{equation}
At a regular configuration where $\overline H_p$ is nonsingular,
\begin{equation}
 \boxed{
 \vect x_p^{(n+1)}
 =-\overline H_p^{-1}
 \left(\overline H_a\vect x_a^{(n+1)}
       +S\widehat{\vect g}_n\right).}
 \label{eq:C27-passive-solution}
\end{equation}
The matrices $\overline H_a$ and $\overline H_p$ are evaluated once at the
current configuration.  They are not differentiated in this construction;
the geometric differentiation has already been accounted for by the Bell
residuals.  Thus a single factorization of $\overline H_p$ serves all requested
orders at that instant.

If the redundant rectangular system is retained, the pseudoinverse formula of
\eqref{eq:C26-rectangular-passive-rates} applies to the right-hand side of
\eqref{eq:C27-passive-recurrence}, together with the corresponding range
compatibility condition at every order.

\subsection{Residuals through order four}

Suppressing the fixed superscript $P$, the first four nonzero branch
residuals are
\begin{align}
 \dvec\rho_{\ell,1}
 &=\sum_p \dot{\dsc\theta}_{\ell p}\dvec V_{\ell p}^{[1]},
 \nonumber\\
 \dvec\rho_{\ell,2}
 &=\sum_p\left(
 2\ddot{\dsc\theta}_{\ell p}\dvec V_{\ell p}^{[1]}
 +\dot{\dsc\theta}_{\ell p}\dvec V_{\ell p}^{[2]}
 \right),
 \nonumber\\
 \dvec\rho_{\ell,3}
 &=\sum_p\left(
 3\dsc\theta_{\ell p}^{(3)}\dvec V_{\ell p}^{[1]}
 +3\ddot{\dsc\theta}_{\ell p}\dvec V_{\ell p}^{[2]}
 +\dot{\dsc\theta}_{\ell p}\dvec V_{\ell p}^{[3]}
 \right),
 \nonumber\\
 \dvec\rho_{\ell,4}
 &=\sum_p\left(
 4\dsc\theta_{\ell p}^{(4)}\dvec V_{\ell p}^{[1]}
 +6\dsc\theta_{\ell p}^{(3)}\dvec V_{\ell p}^{[2]}
 +4\ddot{\dsc\theta}_{\ell p}\dvec V_{\ell p}^{[3]}
 +\dot{\dsc\theta}_{\ell p}\dvec V_{\ell p}^{[4]}
 \right).
 \label{eq:C27-low-order-residuals}
\end{align}
Here $\sum_p$ means $\sum_{p=1}^{m_\ell}$.  Together with
$\dvec\rho_{\ell,0}=\dvec0$, the platform twist jet through geometric order
four requires joint-coordinate derivatives through fifth order.  The
coefficients $1,2,1$; $1,3,3,1$; and
$1,4,6,4,1$ are inherited from the Leibniz separation in the serial formula,
while each $\dvec V^{[s]}$ retains the ordered Bell products inside its branch.

\subsection{Continuation of the \texorpdfstring{$RR+RRR$}{RR+RRR} wrist through order four}

Continue the same mechanism, source trajectory, and instant
$t_0=\pi/(6\Omega)=\pi/6\ \mathrm{s}$ used
in Section~6.  The prescribed active jet, obtained by differentiating
\eqref{eq:C26-wrist-source-trajectory}, is
\begin{equation}
\begin{array}{c|rrrr}
 &\dot q\ [\mathrm{rad/s}]&\ddot q\ [\mathrm{rad/s^2}]
 &q^{(3)}\ [\mathrm{rad/s^3}]&q^{(4)}\ [\mathrm{rad/s^4}]\\ \hline
q_1&1.360350&1.570796&-5.441398&-6.283185\\
q_2&0.785398&-2.720699&-3.141593&10.882796
\end{array}
\label{eq:C27-wrist-active-jet}
\end{equation}
The first passive derivative is already known from
\eqref{eq:C26-wrist-passive-result}.  At each new order $k=2,3,4$, the unknown
$\vect x_p^{(k)}$ is obtained from
\begin{equation}
 \overline H_p\vect x_p^{(k)}
 =-\overline H_a\vect x_a^{(k)}-\vect g_{k-1},
 \label{eq:C27-wrist-order-system}
\end{equation}
using exactly the numerical $\overline H_p$ displayed in
\eqref{eq:C26-wrist-Hp-numeric}.  The changing directions of the three moving
axes are not ignored: all their derivatives are contained in the known Bell
residual $\vect g_{k-1}$.

For this trajectory the successive residual differences, whose units follow
the derivative order of the equation in which they occur, are
\begin{equation}
 \vect g_1=\begin{bmatrix}0.040367\\-1.932536\\0.681028\end{bmatrix}
 \frac{\mathrm{rad}}{\mathrm{s^2}},\quad
 \vect g_2=\begin{bmatrix}2.430547\\9.813511\\6.309067\end{bmatrix}
 \frac{\mathrm{rad}}{\mathrm{s^3}},\quad
 \vect g_3=\begin{bmatrix}27.173195\\66.284112\\11.444111\end{bmatrix}
 \frac{\mathrm{rad}}{\mathrm{s^4}}.
 \label{eq:C27-wrist-residuals}
\end{equation}
They are nonzero already at acceleration order and are assembled only from
the configuration and previously computed derivatives.  Solving the same
$3\times3$ passive system three more times gives
\begin{equation}
\begin{array}{c|rrrr}
 &\dot\phi\ [\mathrm{rad/s}]&\ddot\phi\ [\mathrm{rad/s^2}]
 &\phi^{(3)}\ [\mathrm{rad/s^3}]&\phi^{(4)}\ [\mathrm{rad/s^4}]\\ \hline
\phi_{21}&0.500030&-4.130558&-6.260549&11.575332\\
\phi_{22}&0.937274&1.031338&5.337970&55.143241\\
\phi_{32}&-1.055127&-1.351931&11.961740&36.718826
\end{array}
\label{eq:C27-wrist-passive-jet}
\end{equation}
Thus one
factorization of $\overline H_p$ produces the entire passive jet; only the
right-hand side changes with order.

Finally, reconstruct the platform angular-velocity jet from the $RR$ branch,
or independently from the $RRR$ branch.  With angular components ordered as
$(X,Y,Z)$,
\begin{equation}
\begin{array}{c|c|rrr}
 &&X&Y&Z\\ \hline
\vect\omega&\mathrm{rad/s}&1.360350&-0.191353&0.461967\\
\dot{\vect\omega}&\mathrm{rad/s^2}&1.570796&0.952259&-4.076445\\
\vect\omega^{(2)}&\mathrm{rad/s^3}&-5.441398&12.406824&-2.638865\\
\vect\omega^{(3)}&\mathrm{rad/s^4}&-6.283185&33.286090&47.667852
\end{array}
\label{eq:C27-wrist-platform-jet}
\end{equation}
and all linear components at $O$ vanish.  Substitution of
\eqref{eq:C27-wrist-active-jet}--\eqref{eq:C27-wrist-passive-jet} into
\eqref{eq:C27-wrist-order-system} closes every order to the displayed
precision.  This example realizes the nontrivial map
\begin{equation}
 j^4\vect x_a(t_0)\longmapsto
 \bigl(j^4\vect x_p(t_0),j^3{}^0\vect W_P(t_0)\bigr)
 \label{eq:C27-wrist-input-output-map}
\end{equation}
on the same spatial parallel mechanism used for the first-order calculation.
The geometry and commanded motion are those of \citet[Ch.~9 and
Example~9.3]{Gallardo2016}; the active--passive Bell organization and the
order-four continuation are the present formulation.

The accompanying script \texttt{validate\_wrist.py} provides an independent
symbolic check: it reproduces the passive coordinate derivatives through order
four and the platform angular jet through geometric order three directly from
the closed-form wrist geometry rather than reusing the Bell residual assembly. It
evaluates the same instantaneous $\overline H_p$ throughout and verifies the
first-order closure residual below $10^{-12}$, making the numerical wrist
continuation independently reproducible.

\subsection{Recursive evaluation and closure check}

Given the active jet through order $N+1$, the passive jet is evaluated in
ascending order:
\begin{enumerate}[label=\arabic*.]
 \item evaluate the branch axes, Bell factors, and $\overline H_p$ at the
 current configuration, then factorize $\overline H_p$;
 \item use $n=0$ in \eqref{eq:C27-passive-solution} to obtain passive rates;
 \item for $n=1,\ldots,N$, assemble each $\dvec\rho_{\ell,n}$ only from the
 already known jet, form $\dvec g_n$, and solve
 \eqref{eq:C27-passive-solution};
 \item reconstruct the platform jet from any branch by
 \eqref{eq:C27-highest-derivative-split} and verify all other branches against
 it.
\end{enumerate}
The recurrence is triangular in derivative order.  Its validity requires the
same regularity and compatibility hypotheses as Section~6; Bell polynomials
organize the right-hand sides but do not remove a kinematic singularity.

\section{Affine invariants and higher-order vector fields of the platform}
\label{sec:affine-platform-fields}

Sections~6--7 determine the platform twist and its successive derivatives.
The exact relations recalled at the beginning of this manuscript already
solve the remaining rigid-body problem: the operator polynomial $P_n[D]$ of
\citet{Condurache2022} maps the finite spatial-twist jet directly to the two
affine invariants, and those invariants determine the field at every platform
point.  The purpose of this section is therefore not to derive a second
reconstruction.  It applies that exact map to the platform output of
Section~7 and writes the resulting velocity, acceleration, jerk, and snap
fields explicitly.

Throughout the section all ordinary derivatives are resolved in the same
fixed frame.  In particular,
\begin{equation}
 \widetilde{\vect\omega}=\dot R R^T,
 \label{eq:C28-spatial-angular-convention}
\end{equation}
not $R\dot R^T$.  Thus a body-resolved twist jet must first be transported to
the fixed frame before the formulas below are applied.  The numerical example
below is evaluated directly in its fixed $XYZ$ frame and needs no additional
transformation.

\subsection{Affine field and its invariants}

Let the current position of a material point of the platform be
\begin{equation}
 \vect\rho(t)=R(t)\vect r+\vect p(t),
 \qquad R(t)\in SO(3),
 \label{eq:C28-rigid-motion}
\end{equation}
where $\vect r$ is constant in the platform.  For $n\geq1$, define
\begin{equation}
 \Phi_n:=R^{(n)}R^T,
 \qquad
 \vect a_n:=\vect p^{(n)}-\Phi_n\vect p.
 \label{eq:C28-affine-invariants}
\end{equation}
Then the $n$th material derivative is the affine vector field
\begin{equation}
 \boxed{
 \vect a_{\rho}^{[n]}
 :=\frac{d^n\vect\rho}{dt^n}
 =\vect a_n+\Phi_n\vect\rho.}
 \label{eq:C28-affine-field}
\end{equation}
The pair $(\vect a_n,\Phi_n)$ is independent of the selected point and is
therefore the vector--tensor invariant pair of the order-$n$ field.  The
orders $n=1,2,3,4$ are respectively velocity, acceleration, jerk, and snap.
The units are
\begin{equation}
 [\Phi_n]=\mathrm{s}^{-n},\qquad
 [\vect a_n]=L\,\mathrm{s}^{-n},\qquad
 [\vect a_{\rho}^{[n]}]=L\,\mathrm{s}^{-n};
 \label{eq:C28-field-units}
\end{equation}
radians are dimensionless in these tensor relations.

For the spatial twist
\begin{equation}
 {}^0\vect W_P=(\vect\omega,\vect v),
 \qquad
 \vect v=\dot{\vect p}-\widetilde{\vect\omega}\vect p,
 \label{eq:C28-spatial-twist-components}
\end{equation}
the first invariants are simply
\begin{equation}
 \Phi_1=\widetilde{\vect\omega},
 \qquad \vect a_1=\vect v.
 \label{eq:C28-first-invariants}
\end{equation}

\subsection{Exact operator-polynomial map from the twist jet}

Suppose the twist jet
\begin{equation}
 \left\{(\vect\omega^{(r)},\vect v^{(r)})\right\}_{r=0}^{N-1}
 \label{eq:C28-twist-jet-input}
\end{equation}
is known in the fixed frame.  Let $D=d/dt$.  The symbol $P_n[D]$ denotes the
ordered polynomial map of \citet{Condurache2022}; it is evaluated as a whole,
not composed with a subsequent multiplication operator.  Operationally, we
use the unambiguous recurrence on already evaluated objects
\begin{equation}
 \Phi_1=W_0,\quad \vect a_1=\vect v,\qquad
 \Phi_{n+1}=\dot\Phi_n+\Phi_nW_0,\quad
 \vect a_{n+1}=\dot{\vect a}_n+\Phi_n\vect v,
 \qquad W_0=\widetilde{\vect\omega}.
 \label{eq:C28-exact-P-recurrence}
\end{equation}
The exact relations used here are
\begin{equation}
 \boxed{
 \vect a_n=P_n[D]\vect v,
 \qquad
 \Phi_n=P_n[D]W_0,
 \qquad
 \vect a_\rho^{[n]}=\vect a_n+\Phi_n\vect\rho.}
 \label{eq:C28-exact-twist-to-field-map}
\end{equation}
Thus the requested physical fields follow algebraically from the complete
twist jet; neither the pose nor its derivatives have to be reconstructed.
All numerical values reported below are evaluated from the exact polynomial
map \eqref{eq:C28-exact-twist-to-field-map}; the object recurrence above is an
equivalent implementation and a consistency check.  The twist jet through
$\vect W_P^{(3)}$ is exactly sufficient for the complete snap field.

This map contains only additions and ordered matrix--vector or matrix--matrix
products.  It requires neither numerical differentiation nor the inversion
or solution of a matrix system.  Hence the passage from an already known
twist jet to $(\vect a_n,\Phi_n)$ introduces no kinematic singularity and no
condition-number amplification associated with a linear solve.  In floating
point arithmetic, exact input data are affected only by rounding in these
algebraic operations.  Errors already present in the twist jet are propagated
by the ordered polynomial and may be amplified at high order, for example by
terms scaling as $\|W_0\|^4$ or $\|W_2\|\,\|W_0\|$ in $\Phi_4$.  This remains true
in the neighbourhood of a singular mechanism configuration.  Any loss of
accuracy that may already have occurred while determining the twist jet from
the closure equations belongs to the preceding stage, not to the exact affine
map considered here.

\subsection{Explicit fields through snap}

Write
\begin{equation}
 W_0:=\widetilde{\vect\omega},\quad
 W_1:=\widetilde{\dot{\vect\omega}},\quad
 W_2:=\widetilde{\vect\omega^{(2)}},\quad
 W_3:=\widetilde{\vect\omega^{(3)}}.
 \label{eq:C28-W-notation}
\end{equation}
Direct expansion of \eqref{eq:C28-exact-twist-to-field-map} gives
\begin{align}
 \Phi_1={}&W_0,
 &\Phi_2={}&W_1+W_0^2,
 \nonumber\\
 \Phi_3={}&W_2+2W_1W_0+W_0W_1+W_0^3,
 \nonumber\\
 \Phi_4={}&W_3+3W_2W_0+3W_1^2+3W_1W_0^2+W_0W_2
 \nonumber\\
 &{}+2W_0W_1W_0+W_0^2W_1+W_0^4,
 \label{eq:C28-Phi-through-four}
\end{align}
\begin{align}
 \vect a_1={}&\vect v,
 &\vect a_2={}&\dot{\vect v}+W_0\vect v,
 \nonumber\\
 \vect a_3={}&\vect v^{(2)}+2W_1\vect v+W_0\dot{\vect v}+W_0^2\vect v,
 \nonumber\\
 \vect a_4={}&\vect v^{(3)}+3W_2\vect v+3W_1\dot{\vect v}
 +3W_1W_0\vect v+W_0\vect v^{(2)}
 +2W_0W_1\vect v+W_0^2\dot{\vect v}+W_0^3\vect v.
 \label{eq:C28-a-through-four}
\end{align}
Factor order is essential.  No commutation of $W_iW_j$ has been assumed.
Substitution in \eqref{eq:C28-affine-field} yields the four physical vector
fields:
\begin{equation}
 \begin{array}{c|c}
 n&\text{platform field at the current point }\vect\rho\\ \hline
 1&\vect v_{\rho}=\vect a_1+\Phi_1\vect\rho\\
 2&\vect a_{\rho}=\vect a_2+\Phi_2\vect\rho\\
 3&\vect j_{\rho}=\vect a_3+\Phi_3\vect\rho\\
 4&\vect s_{\rho}=\vect a_4+\Phi_4\vect\rho
 \end{array}
 \label{eq:C28-four-fields}
\end{equation}

\subsection{Spatial \texorpdfstring{$6$-RUS}{6-RUS} configuration with active revolutes}

The spherical wrist is replaced by the Hunt-type $6$-RUS configuration of
\citet{Gil2004}.  Each branch has an actuated base revolute followed by passive
universal and spherical joints.  The fixed actuator axes form an equilateral
triangle of side $1\,\mathrm m$; adjacent axis points on an edge are separated
by $0.1\,\mathrm m$.  All crank and rod lengths are, respectively,
$0.1\,\mathrm m$ and $0.6\,\mathrm m$, and the moving platform is an
equilateral triangle of side $0.5\,\mathrm m$.  At the evaluation instant,
\begin{equation}
 \theta_i=30^\circ,\qquad
 \dot{\boldsymbol\theta}=(10,-5,-5,5,-5,10)^T\ \mathrm{rad/s},
 \qquad \boldsymbol\theta^{(r)}=\vect0\quad(r=2,3,4).
 \label{eq:C28-6rus-input-jet}
\end{equation}
Only the architecture, geometry, configuration, and active input jet are
taken from the source.  The platform jet and all affine-field values below
are recomputed internally from the unrounded analytic geometry; the rounded
output values printed by the source are not used as reference data.

The three platform vertices at this instant are
\begin{equation}
\begin{aligned}
 \vect x_{123}&=(-0.144337567,0.25,0.612731434)^T\ \mathrm m,\\
 \vect x_{145}&=(0.288675135,0,0.612731434)^T\ \mathrm m,\\
 \vect x_{161}&=(-0.144337567,-0.25,0.612731434)^T\ \mathrm m.
\end{aligned}
\label{eq:C28-6rus-platform-points}
\end{equation}
Algebraic differentiation of the six exact rod-length constraints yields the
spatial twist jet
\begin{equation}
\begin{array}{c|rrr|rrr}
r&\multicolumn{3}{c|}{\vect\omega^{(r)}\ [\mathrm{s}^{-(r+1)}]}&
\multicolumn{3}{c}{\vect v^{(r)}\ [\mathrm{m\,s}^{-(r+1)}]}\\
 &X&Y&Z&X&Y&Z\\ \hline
0&-1.538928&0.470383&-1.764660&-0.288218&-1.961777&0.135788\\
1& 9.222203&8.726590&0&-3.303146&5.650734&-6.907909\\
2&90.220744&-131.007308&163.979476&72.122563&176.100766&-6.994190\\
3&-26.845951&-3753.418362&235.301173&1772.967751&-300.641902&1358.234914
\end{array}
\label{eq:C28-6rus-twist-jet}
\end{equation}
Application of the exact map \eqref{eq:C28-exact-twist-to-field-map} gives
nonzero vector invariants at every order:
\begin{equation}
\begin{array}{c|rrr}
n&\multicolumn{3}{c}{\vect a_n\ [\mathrm{m\,s}^{-n}]}\\
 &X&Y&Z\\ \hline
1&-0.288218&-1.961777&0.135788\\
2&-6.701143&6.368309&-3.753304\\
3&83.964904&179.645418&-44.795937\\
4&2882.354753&-592.446164&749.646950
\end{array}
\label{eq:C28-6rus-vector-invariants}
\end{equation}
The corresponding tensor invariants are
\begin{align}
\Phi_1={}&\begin{bmatrix}
0&1.764660&0.470383\\-1.764660&0&1.538928\\-0.470383&-1.538928&0
\end{bmatrix}\ \mathrm{s}^{-1},\nonumber\\
\Phi_2={}&\begin{bmatrix}
-3.335286&-0.723885&11.442274\\-0.723885&-5.482325&-10.052269\\
-6.010905&8.392138&-2.589559
\end{bmatrix}\ \mathrm{s}^{-2},\nonumber\\
\Phi_3={}&\begin{bmatrix}
-12.314507&-196.565585&-149.964231\\169.290702&42.576915&-114.397615\\
101.142062&68.199216&30.262408
\end{bmatrix}\ \mathrm{s}^{-3},\nonumber\\
\Phi_4={}&\begin{bmatrix}
1241.926843&696.448934&-4765.908160\\706.298573&1504.277031&641.097103\\
3088.347393&560.916266&270.322229
\end{bmatrix}\ \mathrm{s}^{-4}.
\label{eq:C28-6rus-tensor-invariants}
\end{align}
For the vertex $\vect x_{123}$, the complete physical fields are
\begin{equation}
\begin{array}{c|c|rrr}
\text{field}&\text{unit}&X&Y&Z\\ \hline
\dot{\vect x}_{123}&\mathrm{m/s}&0.441165&-0.764120&-0.181050\\
\ddot{\vect x}_{123}&\mathrm{m/s^2}&0.610334&-1.057130&-2.374374\\
\vect x_{123}^{(3)}&\mathrm{m/s^3}&-55.286845&95.759624&-23.802004\\
\vect x_{123}^{(4)}&\mathrm{m/s^4}&-43.011454&74.498023&609.746394
\end{array}
\label{eq:C28-6rus-point-fields}
\end{equation}

The computational route used for this configuration is deliberately
independent of the branchwise Bell recurrence of Sections~4--7: it
differentiates the rod constraints by truncated Taylor arithmetic in pose
coordinates.  It therefore tests closure consistency and the twist-to-field
stage without reusing their derivation. It does not reconstruct the passive
coordinates of the universal and spherical joints and therefore is not, by
itself, an end-to-end implementation of the branchwise Bell recurrence. The
wrist and the generic $3C$ test exercise the closure and serial Bell machinery,
respectively; the three tests are complementary.

Two independent internal paths are compared.  The first differentiates
$\vect p+R\vect r_j$ directly; the second evaluates
$\vect a_n+\Phi_n\vect x_j$ from the exact operator-polynomial map.  Over the
three vertices and an auxiliary platform point with local coordinates
$(0,0,1/6)^T\,\mathrm m$, chosen off their plane to test full spatial rigidity,
the maximum absolute residual at every order from one through four is
\begin{equation}
 \max_{1\le n\le4}\varepsilon_n<10^{-12},
 \label{eq:C28-6rus-internal-field-residuals}
\end{equation}
in the corresponding SI units.  The Taylor coefficients of all six closure
equations and the coefficientwise orthogonality conditions of $R^TR-I$ also
vanish through order four below $10^{-12}$.  These are environment-robust
floating-point bounds; exact machine-level values remain in the JSON output.
They are internal checks;
they are not comparisons against rounded published outputs.

\subsection{Computational map and checks}

The complete continuation of Sections~6--7 is now
\begin{equation}
 j^4\vect x_a
 \longmapsto
 \left(j^4\vect x_p,j^3{}^0\vect W_P\right)
 \longmapsto
 \left\{(\vect a_n,\Phi_n)\right\}_{n=1}^{4}
 \longmapsto
 \left\{\vect a_{\rho}^{[n]}\right\}_{n=1}^{4}.
 \label{eq:C28-complete-map}
\end{equation}
The following checks are inexpensive and should accompany every numerical
implementation:
\begin{enumerate}[label=\arabic*.]
 \item verify $\Phi_1+\Phi_1^T=0$;
 \item verify factor order in every direct expansion of $P_n[D]$;
 \item verify the dimensions in \eqref{eq:C28-field-units};
 \item evaluate the same point field once from
 $d^n(R\vect r+\vect p)/dt^n$ and once from
 $\vect a_n+\Phi_n\vect\rho$;
 \item verify all branch closures and $R^TR=I$ through the requested order.
\end{enumerate}
Thus the twist jet is not the final physical output.  It is the finite input
needed to reconstruct, point-independently, the affine invariants and then the
velocity, acceleration, jerk, and snap at any requested point of the moving
platform.

\section{Conclusions}
\label{sec:conclusions}

This paper has established a continuous kinematic route from joint-coordinate
jets to the higher-order motion fields of a parallel mechanism's moving
platform.  The construction separates three operations that are often mixed:
serial propagation along each branch, active--passive closure across the
branches, and reconstruction of the affine fields over the rigid platform.
Keeping these stages distinct makes both the algebra and the numerical checks
explicit.

At the serial level, a cylindrical joint was preserved as one native physical
block.  Its coaxial rotation and translation were encoded by one dual scalar,
with revolute and prismatic joints recovered by restriction.  Because the
derivatives of a single fixed-axis generator commute, ordinary Bell
polynomials provide the joint-level coefficients.  Noncommutativity between
different joints is retained through the ordered chain product. The resulting
initial-frame and terminal-resolved covariant constructions resolve the same
twist jet in the initial and terminal frames, respectively, and therefore supply the branch descriptions
needed for a common platform closure.

At the parallel level, repeated differentiation of the closure equations was
organized by the Leibniz rule and joint-level Bell factors, consistently with
the multivariate Fa\`a di Bruno structure. At every order, the unknown
highest passive derivative enters linearly, whereas all nonlinear terms depend
only on already available lower-order jets.  Consequently, one factorization
of the passive closure Jacobian can be reused through the requested derivative
order at a fixed regular configuration.  This triangular structure is not a
regularization method: if the passive Jacobian loses rank, the selected local
parameterization and its unique jet continuation may fail.

The closed platform twist jet was finally converted by the exact
operator-polynomial map into the point-independent pairs
$(\vect a_n,\Phi_n)$.  These pairs generate the velocity, acceleration, jerk,
and snap at any material point by the affine relation
\begin{equation}
 \vect a_{\rho}^{[n]}=\vect a_n+\Phi_n\vect\rho,
 \qquad n=1,\ldots,4.
 \label{eq:C29-final-affine-field}
\end{equation}
No additional mechanism solve is required at this stage; all conditioning
associated with the active--passive partition has already entered through the
closure problem.

Three complementary numerical checks served different purposes. A generic noncoplanar $3C$
chain with nonzero rotational and translational coordinates tested the ordered
initial/terminal covariant construction directly. The $RR+RRR$ spherical wrist exposed
the closure recurrence in a compact setting. The Hunt-type $6$-RUS
configuration then tested an independently reconstructed spatial platform jet,
with six active revolute joints and nonzero translational as well as rotational
platform fields, together with the exact twist-to-field map. This
constraint-based test does not reconstruct the passive universal and spherical
coordinates and is therefore complementary to, rather than a replacement for,
the branchwise Bell tests. The numerical values were recomputed from the mechanism geometry and
input jet rather than fitted to rounded published outputs.  Direct
differentiation of the rigid motion and independent evaluation of the affine
fields agreed through snap with residuals below $10^{-12}$ in the corresponding
SI units; all six branch closures vanished through fourth order below the same
threshold.  Rotation orthogonality and platform rigidity supplied
additional internal checks.

The present results are limited to kinematics with fixed topology and a
regular active--passive partition at the evaluation configuration.  They do
not address singular continuation, branch switching, dynamics, compliance, or
contact.  Natural extensions are therefore the treatment of rank-deficient
closure using explicitly stated generalized-coordinate choices, conditioning
and error propagation for high-order jets, and integration of the resulting
affine fields into higher-order inverse dynamics, feedforward control, and
jerk- or snap-constrained trajectory generation.  Within its stated regular
domain, however, the Bell construction provides an order-consistent and
internally verifiable framework from native lower-pair joints to complete
rigid-platform motion fields.

\bibliographystyle{unsrtnat}
\bibliography{references}
\end{document}